\documentclass[11pt,a4paper]{article}
\usepackage[hyperref]{naaclhlt2019}
\usepackage{times}
\usepackage{latexsym}

\usepackage{url}

\aclfinalcopy % Uncomment this line for the final submission
\def\aclpaperid{1257} %  Enter the acl Paper ID here

\usepackage{amsmath}
\usepackage{amsfonts}

\usepackage{blkarray}
\usepackage[utf8]{inputenc}
\usepackage{microtype}
\usepackage[safe]{tipa}
\usepackage{booktabs}
\usepackage{multirow}
\usepackage{multicol}
\usepackage{graphicx}
\usepackage{xcolor}

\usepackage{cleveref}
\crefname{section}{\S}{\S\S}
\Crefname{section}{\S}{\S\S}
\crefname{table}{Table}{}
\crefname{figure}{Figure}{}
\crefname{algorithm}{Algorithm}{}
\crefname{equation}{eq.}{}
\crefname{appendix}{App.}{}
\crefformat{section}{\S#2#1#3}  % remove space between section symbol and the number

\newcommand\bluebox[1]{%
  \colorbox{blue!30}{$#1$}%
}
\newcommand\redbox[1]{%
  \colorbox{red!30}{$#1$}%
}

\newcommand\greenbox[1]{%
  \colorbox{green!30}{$#1$}%
}

\newcommand{\word}[1]{\textit{#1}}
\renewcommand{\vec}[1]{\mathbf{#1}}
\renewcommand{\bold}[1]{{\boldsymbol #1}}
\newcommand{\langembed}{{\boldsymbol \lambda}}
\newcommand{\paramembed}{\vec{e}_{\pi_i}}
\newcommand{\R}{\mathbb{R}}
\newcommand{\sigmoid}{\textit{sigmoid}}

\title{A Probabilistic Generative Model of Linguistic Typology}
 
\author{Johannes Bjerva$^{\textrm{\normalfont\textschwa}}$ \text{ }  Yova Kementchedjhieva$^{\textrm{\normalfont \textschwa}}$ \text{ }  Ryan Cotterell$^{\textrm{\normalfont \textipa{P},\textipa{H}}}$ \text{ } Isabelle Augenstein$^{\textrm{\normalfont \textschwa}}$ \\
${}^{\textrm{\textschwa}}$Department of Computer Science, University of Copenhagen \\
${}^{\textrm{\textipa{P}}}$Department of Computer Science, Johns Hopkins University \\
${}^{\textrm{\textipa{H}}}$Department of Computer Science and Technology, University of Cambridge  \\
{\tt {bjerva,yova,augenstein}@di.ku.dk, rdc42@cam.ac.uk}
}

\date{}

\begin{document}
\maketitle
\begin{abstract}
In the principles-and-parameters framework, the structural features of languages depend on parameters that may be toggled on or off, with a single parameter often dictating the status of multiple features. 
The implied covariance between features inspires our probabilisation of this line of linguistic inquiry---we develop a generative model of language based on exponential-family matrix factorisation.
By modelling all languages and features within the same architecture, we show how structural similarities between languages can be exploited to predict typological features with near-perfect accuracy, outperforming several baselines on the task of predicting held-out features.
Furthermore, we show that language embeddings pre-trained on monolingual text allow for generalisation to unobserved languages. 
This finding has clear practical and also theoretical implications: the results confirm what linguists have hypothesised, i.e.~that there are significant correlations between typological features and languages.
\end{abstract}

\section{Introduction}
Linguistic typologists dissect and analyse languages in terms of their
structural properties \citep{croft2002typology}. For instance, consider the phonological
property of word-final obstruent decoding: German devoices word-final
obstruents (\word{Zug} is pronounced /zuk/), whereas English does not
(\word{dog} is pronounced /d\textturnscripta g/). In the tradition of
generative linguistics, one line of typological analysis is the
principles-and-parameters framework \cite{chomsky1981lectures}, which posits the existence of a set of universal parameters, switches as it were, that languages
toggle. One arrives at a kind of factorial typology, to borrow
terminology from optimality theory \cite{prince2008optimality}, through different
settings of the parameters. Within the principle-and-parameters research
program, then, the goal is to identify the parameters that serve
as axes, along which languages may vary.

\begin{figure}
  \centering
  \includegraphics[width=\columnwidth]{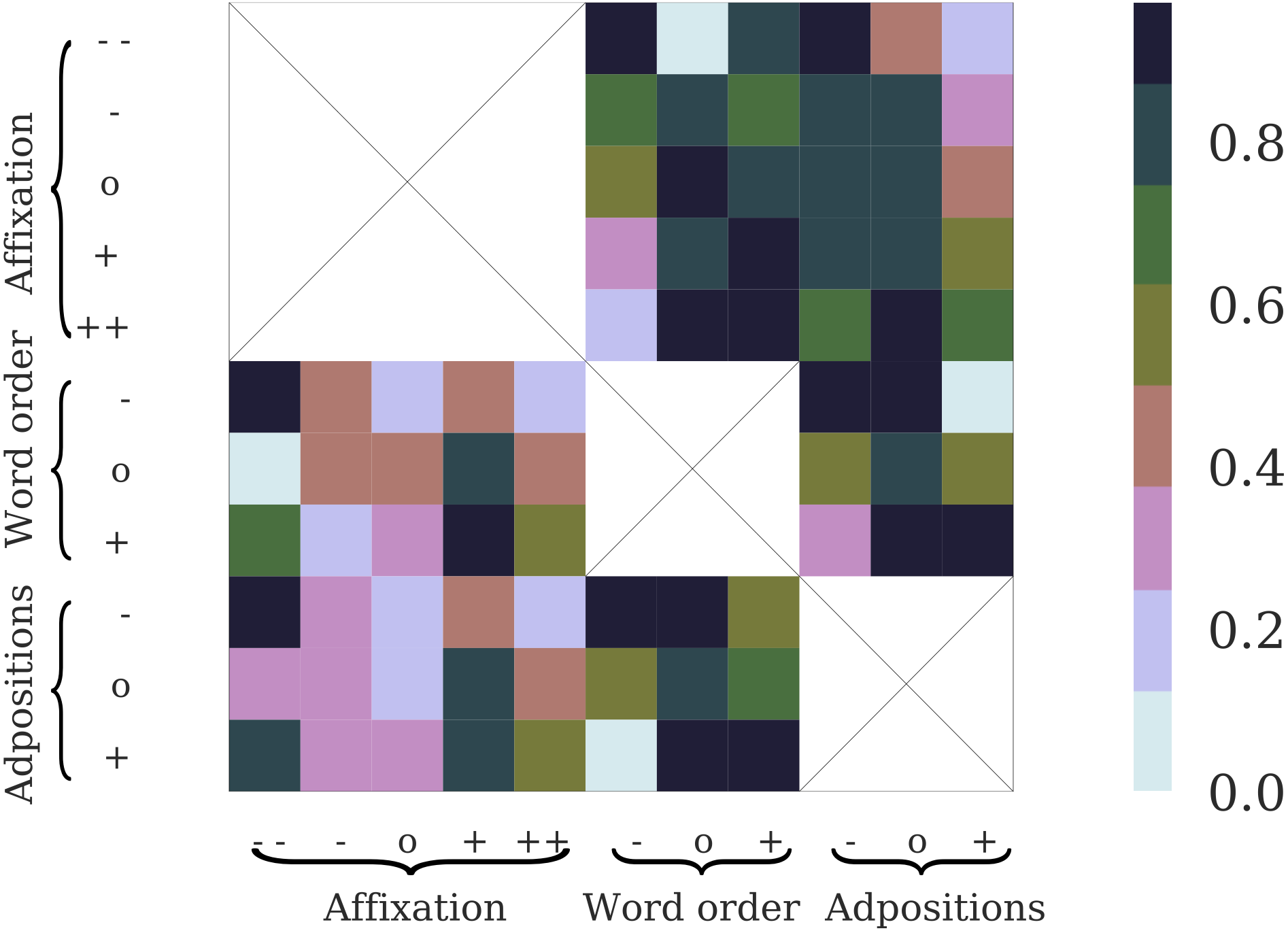}
  \caption{Correlations between selected typological parameters. Feature values are classified according to head-directionality (head-initial \textbf{+}, head-final \textbf{-}, no dominant order \textbf{o}). For instance, \textbf{++} under Affixation means strongly suffixing.}
  \label{fig:corr-heat} 
\end{figure}

It is not enough, however, to simply write down the set of parameters
available to language. Indeed, one of the most interesting facets of
typology is that different parameters are correlated. To illustrate this
point, we show a heatmap in \cref{fig:corr-heat} that shows the correlation
between the values of selected parameters taken from a typological knowledge base (KB). Notice how head-final word order, for example, highly correlates with strong suffixation. The primary contribution 
of this work is a probabilisation of typology inspired by the
principles-and-parameters framework. We assume a given set of
typological parameters and develop a generative model of a language's
parameters, casting the problem as a form of
exponential-family matrix factorisation. We observe a binary matrix
that encodes the settings of each parameter for each language. For
example, the Manchu head-final entry of this matrix would be set to 1, because
Manchu is a head-final language.  The goal of our model is to
\emph{explain} each entry of matrix as arising through the dot product
of a language embedding and a parameter embedding passed through a
sigmoid.

We test our model on The World Atlas of Language Structures (WALS), the largest available knowledge base of typological parameters at the lexical, phonological, syntactic and semantic level. 
Our contributions are:
(i) We develop a probabilisation of typology inspired by the principles-and-parameters framework. 
(ii) We introduce the novel task of typological collaborative filtering, where we observe some of a
language's parameters, but hold some out. At evaluation time, we predict the held-out parameters using the generative model.
(iii) We develop a semi-supervised extension, in which we incorporate language embeddings output by a neural language model, thus improving performance with unlabelled data. Indeed, when we partially observe
some of the typological parameters of a language, we achieve near-perfect ($97\%$) accuracy on the prediction of held-out parameters. 
(iv) We perform an extensive qualitative and quantitative analysis of our method.\looseness=-1

\section{Typology in Generative Grammar}\label{sec:typology}
What we will present in this paper is a generative model that corresponds
to a generative tradition of research in linguistic typology. We first outline
the technical linguistic background necessary for the model's exposition.
Chomsky famously argued that the human brain contains a
prior, as
it were, over possible linguistic structures,  which he termed
\textbf{universal grammar} \cite{chomsky1965}.  The connection between Chomsky's Universal Grammar and
  the Bayesian prior is an intuitive one, but the earliest citation we
  know for the connection is \newcite[\S 2]{eisner-2002-cogsci}. As a theory, universal grammar
holds great promise in explaining the typological variation of human
language.  Cross-linguistic similarities and differences may be
explained by the influence universal grammar exerts over language
acquisition and change. While universal grammar arose early on in the
writtings of Chomsky, early work in generative grammar
focused primarily on English \cite{harris1995linguistics}.  Indeed, Chomsky's
\textit{Syntactic Structures} contains exclusively examples in English
\cite{chomsky2002syntactic}.  As the generative grammarians turned their focus to
a wider selection of languages, the \textbf{principles and parameters}
framework for syntactic analysis rose to prominence.  Given the tight
relationship between the theory of universal grammar and typology, principles
and parameters offers a fruitful manner in which to research
typological variation.\looseness=-1

The principles and parameters takes a parametric view of linguistic
typology. The structure of human language is governed by a series of
\textbf{principles}, which are hard constraints on human language. A
common example of a principle is the requirement that every sentence
has a subject, even if one that is not pronounced; see the discussion on
the pro-drop parameter in \newcite{carnie2013syntax}. Principles are
universally true for all languages. On the other hand, languages are
also governed by \textbf{parameters}. Unlike principles, parameters
are the parts of linguistic structure that are allowed to vary. It is
useful to think of parameters as attributes that can take on a variety
of values. As \newcite{chomsky_smith_2000} himself writes ``we can
think of the initial state of the faculty of language as a fixed
network connected to a switch box; the network is constituted of the
principles of language, while the switches are the options to be
determined by experience. When the switches are set one way, we have
Swahili; when they are set another way, we have Japanese. Each
possible human language is identified as a particular setting of the
switches-a setting of parameters, in technical terminology.''

What are possible parameters? Here, in our formalisation of the
parameter aspect of the principles-and-parameters framework, we
take a catholic view of parameters, encompassing all areas of
linguistics, rather than just syntax.  For example, as we saw before,
consider the switch of devoicing word-final obstruents a parameter. 
We note that while principle-and-parameters typology has primarily been applied to
syntax, there are also interesting applications to non-syntactic domains. For instance,
\newcite{van2015parametric} applies a parametric
approach to metrical stress in phonology; this is in line with our view.
In the field of linguistic typology, there is a vibrant line of
research, which fits into the tradition of viewing typological parameters through the lens of principles and parameters. Indeed,
while earlier work due to Chomsky focused on what have come to be called
macro-parameters, many linguists now focus on micro-parameters, which are very
close to the features found in the WALS dataset that we will be modelling \citep{baker,nicolis,biberauer}.
This justifies our viewing WALS through the lens of principles and parameters, even though the authors of WALS adher to the functional-typological school.\footnote{For an overview of differences between these schools, we refer the reader to \citet{haspelmath:2008}.}

Notationally,
we will represent the parameters as a vector $\bold{\pi} =
\left[\pi_1, \ldots, \pi_n\right]$. Each
typological parameter $\pi_i$ is a binary variable; for instance, does
the language admit word-final voiced obstruents?

\begin{figure}
{\small
\[
\begin{blockarray}{l|cccccc}
\ell & \text{SOV} & \cdots & \text{SVO} & \text{Str. Pref.} & \cdots & \text{Str. Suff.} \\
\begin{block}{l[cccccc]}
 \text{eng} & \bluebox{1} & \bluebox{0} & \bluebox{0} & \redbox{0} & \redbox{0} & \redbox{1} \\
 \text{nld} & \redbox{0} & \redbox{1} & \redbox{0} & \bluebox{0} & \bluebox{0} & \bluebox{1} \\
 \text{deu} & \redbox{0} & \redbox{1} & \redbox{0} & \redbox{0} & \redbox{0} & \redbox{1} \\
 \cdots    & \cdots &&& \cdots && \\
 \text{vie} & \bluebox{0} & \bluebox{0} & \bluebox{1} & \bluebox{0} & \bluebox{1} & \bluebox{0} \\
 \text{tur} & \bluebox{0} & \bluebox{1} & \bluebox{0} & \bluebox{0} & \bluebox{0} & \bluebox{1} \\
 \text{mrd} & \bluebox{1} & \bluebox{0} & \bluebox{0} & \greenbox{-} & \greenbox{-} & \greenbox{-} \\
\end{block}
\end{blockarray}
 \]}
\caption{Example of training and evaluation portions of a feature matrix, with word order features (81A, $\text{SOV},\cdots,\text{SVO}$) and affixation features (26A, $\text{Strongly Prefixing},\cdots,\text{Strongly Suffixing}$). We train on the examples highlighted in blue, evaluate on the examples highlighted in red, and ignore those which are not covered in WALS highlighted in green.}
\label{fig:matrix}
\end{figure}

\section{A Generative Model of Typology}\label{sec:model}
We now seek a probabilistic formalisation of the linguistic theory
presented in \cref{sec:typology};
specifically, for every language $\ell$, we seek to explain the observed binary vector of parameters $\bold{\pi}^\ell$:
$\pi^\ell_i = 1$ indicates that the $i^\text{th}$ parameter is ``on''
in language $\ell$. The heart of
our model will be quite simple: every language $\ell$ will have a
language embedding $\langembed^\ell \in \R^d$ and every parameter will have a parameter embedding
$\paramembed \in \R^d$. Now, $\pi^\ell_i \sim \sigmoid(\paramembed^{\top}\langembed^\ell)$. 

This model also takes inspiration from work
in relation extraction \cite{Riedel2013}.
Writing the joint distribution over the entire binary vector of parameters,
we arrive at
\begin{align}
  p(\bold{\pi}^\ell \mid \bold{\lambda}^\ell) &= \prod_{i=1}^{|\bold{\pi}|} p(\pi^\ell_i \mid \bold{\lambda}^\ell) \\
  &= \prod_{i=1}^{|\bold{\pi}|} \textit{sigmoid}\left( \vec{e}_{\pi_i}^{\top} \bold{\lambda}^\ell \right) \\
  &= \prod_{i=1}^{|\bold{\pi}|} \frac{1}{1 + \exp(-\vec{e}_{\pi_i}^{\top} \bold{\lambda}^\ell)} 
\end{align}
We define the the prior over language embeddings:
\begin{equation}\label{eq:prior}
  p(\langembed^\ell) = {\cal N}\left(\bold{0}; \sigma^2 I\right)
\end{equation}
where $\bold{\mu}$ is the mean of the Gaussian whose covariance is fixed at $I$. 
Now, give a collection of languages ${\cal L}$, we arrive at the joint
distribution
\begin{equation}\label{eq:model}
  \prod_{\ell \in {\cal L}} p(\bold{\pi}^{\ell}, \langembed^{\ell}) = \prod_{\ell \in {\cal L}} p(\bold{\pi}^{\ell} \mid \langembed^{\ell} )\cdot p(\langembed^{\ell})
\end{equation}
Note that $p(\langembed^\ell)$ is, spiritually at least, a universal grammar: it is the
prior over what sort of languages can exist, albeit encoded as a real
vector. In the parlance of principles and parameters, the prior represents
the principles.

Then our model parameters are $\bold{\Theta} =
\{\vec{e}_{\pi_1}, \ldots, \vec{e}_{\pi_{|\bold{\pi}|}}, \langembed^1 , \ldots, \langembed^{|\mathcal{L}|}\}$.
Note that for the remainder of the paper, we will never
shorten `model parameters' to simply `parameters' to avoid ambiguity.
We will, however, refer to `typological parameters' as simply `parameters.'

We can view this model as a form of exponential-family matrix
factorisation \cite{DBLP:conf/nips/CollinsDS01}. Specifically, our model seeks to explain a
binary matrix of parameters. We consider such matrices as the one %presented
in \cref{fig:matrix}, which depicts some of the binarised feature values for word order and affixation for English, Dutch, German, Vietnamese, Turkish, and Marind.
We will have some parameters %which are 
seen during training (highlighted in blue), some we use for evaluation (highlighted in red), and some which are unknown due to the nature of WALS (highlighted in green).
Crucially, the model in \cref{eq:model} allows us to learn the correlations between typological parameters, as illustrated in \cref{fig:corr-heat}. 
We train the model over 10 epochs with a batch size of 64, using the Adam optimiser \citep{adam} and $L_2$ regularisation (0.1), which corresponds to the Gaussian prior with variance $\sigma^2 = 10$.

\section{A Semi-Supervised Extension}
\begin{figure}
	\centering
	\includegraphics[width=\columnwidth]{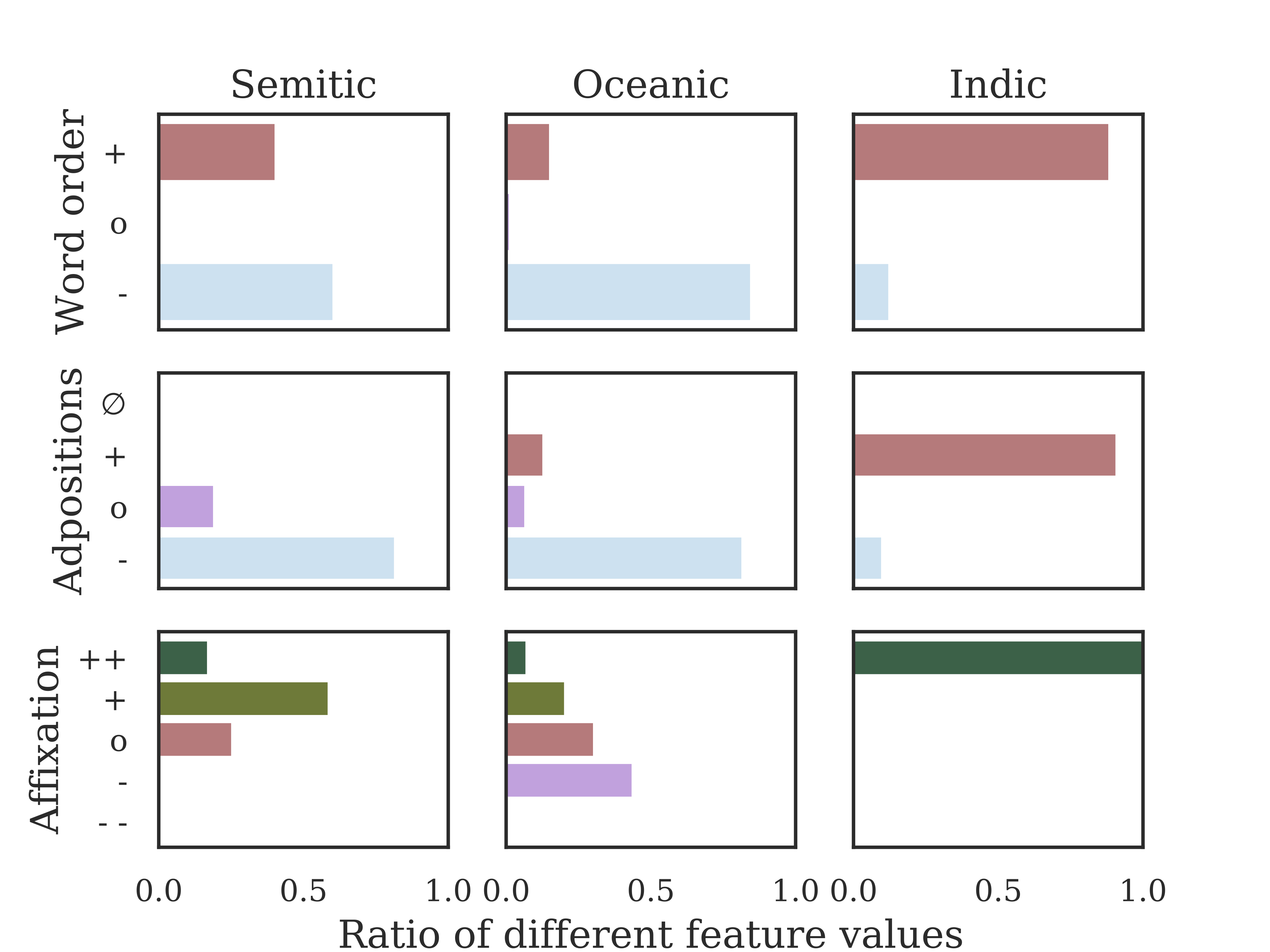}
	\caption{\label{fig:corr-langs}Distribution of feature values across three of the biggest language families in WALS. The representation of feature values is described in Figure~\ref{fig:corr-heat}.}
\end{figure}

A natural question we might ask is if our model can exploit unlabelled monolingual data to
improve its performance. We explain how we can induce language embeddings
from unlabelled data below %in the following section 
and then incorporate
these into our model through the prior \cref{eq:prior}.
This results in a semi-supervised model, as we incorporate an unsupervised pre-training step. This is motivated by the fact that related languages tend to exhibit correlations between each other.
\cref{fig:corr-langs} shows the distribution of a few features within the Semitic, Oceanic, and Indic language branches. Notice, for instance, the skewed distribution of feature values within the Indic branch: languages in that branch are almost exclusively head-initial with respect to word order, order of adposition and noun, and affixation. 

\subsection{Distributional Language Embeddings}
Words can be represented by distributed word representations, currently often in the form of word embeddings.
Similarly to how words can be embedded, so can languages, by associating each language with a real-valued vector known as a \textit{language embedding}.
Training such representations as a part of a multilingual model allows us to infer similarities between languages.
This is due to the fact that in order for multilingual parameter sharing to be successful in this setting, the neural network needs to use the language embeddings to encode features of the languages.
Previous work has explored this type of representation learning in various tasks, such as NMT \citep{malaviya:2017}, language modelling \citep{tsvetkov:2016,ostling_tiedemann:2017}, and tasks representing morphological, phonological, and syntactic linguistic levels \citep{bjerva_augenstein:naacl:18}.

In the context of computational typology, representations obtained through language modelling 
have been the most successful \citep{ostling_tiedemann:2017}.
This approach is particularly interesting since unlabelled data is available for a large portion of the world's languages, meaning that high quality language embeddings can be obtained for more than 1,000 of the world's languages.

\begin{figure}%[!h]%[tbp]
	\centering
	\includegraphics[width=0.7\columnwidth]{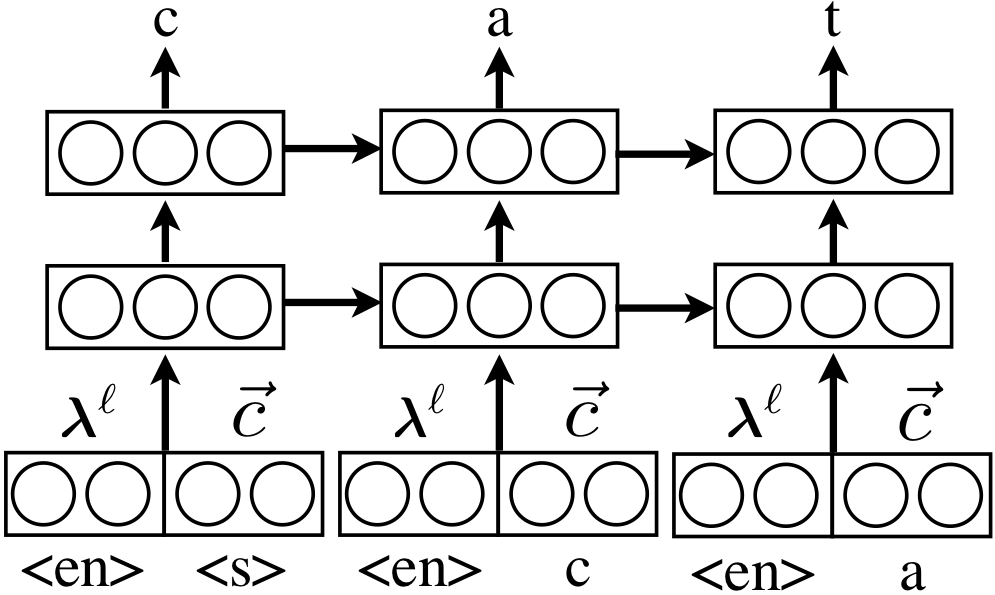}
	\caption{\label{fig:language_model}Char LSTM LM with language embeddings concatenated to char embeddings at every time step.}
\end{figure}

\subsection{Language Embeddings through LMs}
In this work, we use a language modelling objective to pre-train
language embeddings; we train a
character-level neural language model with a distributed language
embedding of language $\ell$.  Specifically, we use the model of
\newcite{ostling_tiedemann:2017}, visualised in
\cref{fig:language_model}.  The model is a stacked character-based
LSTM \cite{lstm} with two layers, followed by a softmax output layer.
In order to accommodate the language embeddings, this relatively
standard model is modified such that language embeddings are
concatenated to the character embeddings at each time step. This method returns a language embedding, which we denote $\tilde{\langembed}^{\ell}$ to distinguish it from the language embeddings $\langembed^{\ell}$ discussed in the previous section.  We use
the same hyperparameter settings as \newcite{ostling_tiedemann:2017},
with 1024-dimensional LSTMs, 128-dimensional character embeddings, and
64-dimensional language embeddings.  Training is done with Adam
\cite{adam}, and using early stopping.

In the semi-supervised regime, we use the estimated
language embedding $\tilde{\langembed}^{\ell}$ from the language model and define the model as follows
\begin{equation}
    p({\boldsymbol {\pi}}^\ell \mid \tilde{\langembed}^\ell) = \prod_{i=1}^{|{\boldsymbol \pi}|} \textit{sigmoid}(\paramembed^{\top} \tilde{{\boldsymbol {\lambda}}}^\ell)
\end{equation}
omitting the learned language embedding $\langembed$ in the matrix factorisation. The likelihood of this model is now convex in the parameter embeddings. In contrast to the full matrix factorisation setting, here, all language-specific knowledge must come from an external source, namely, the unlabelled text. 

\section{A Novel Task: Typological Collaborative Filtering}
In this section, we introduce a novel task for linguistic typology,
which we term \textbf{typological collaborative filtering}. 
Typological KBs such as WALS are known
to be incomplete. In other words, not all parameters $\pi^\ell_i$
are observed for all languages. Thus, a natural question we may ask 
how well our models can predict unobserved (or held-out) parameters.
We view this problem as analogous to collaborative filtering (CF).
Our task is similar to knowledge base population or completion in that we start off with a partly populated KB which we aim to complete, but differs in the aspect that we attempt to infer values from correlations between features and similarities between languages, rather than inferring these from a collection of texts.

CF is a common technique for recommender systems (see §\ref{sec:RelWork}). Consider the task of recommending
a new movie to a customer. Given which movies different users
have liked (equivalent to a typological parameter being `on') and
which movies the user has disliked (equivalent to the typological parameter
being `off'), a CF model tries to figure out the (latent)
preferences of each user and the latent genre of each movie. In our setting,
the languages are analogous to users and movies are analogous to parameters.
Our model in \cref{eq:model} seeks to learn what latent properties of languages
and what latent properties of parameters explain their correlations.

\section{Data}\label{sec:data}
\paragraph{WALS.}
The World Atlas of Language Structures (WALS) is a large knowledge base of typological properties at the lexical, phonological, syntactic and semantic
level on which we will run our experiments.
The documentation of linguistic structure is spread throughout a wide variety of academic works, ranging from field linguistics to grammars describing the nuances of individual grammatical uses.
KB creation is a laborious tasks as it involves distilling knowledge into a single, standardised resource, which, naturally, will be incomplete, prompting the need for methods to complete them \cite{conf/naacl/MinGW0G13}. In the case of WALS, few languages have all values annotated for all of the properties. 
In this section, we offer a formalisation of typological KBs to allow for our development of a probabilistic model over vectors of properties.
 WALS, for instance, contains $n=202$ different parameters
\cite{wals}.

\paragraph{Binarisation of WALS.}
Many common typological KBs, including WALS, the one studied here,
contain binary as well as non-binary parameters. To deal with this, we binarise the KB as follows: Whenever there is a typological parameter that takes $\geq 3$ values, e.g., `Feature 81A: Order of Subject, Object and Verb' which takes the 7 values `SOV', `SVO', `VSO', `VOS', `OVS', `OSV', `No dominant order', we introduce that many binary parameters.
At test time, we get non-binary predictions by using a simple decoder that returns the arg max over the predicted probabilities for the binary features.\looseness=-1

As each typological parameter with $n\geq3$ feature values is coded as a one-hot binary vector of length $n$, we need to make sure that we do not mix a single typological parameter for a language into the training and test sets.
This is visualised in \cref{fig:matrix}, where we train on the blue squares, i.e., the binarised 81A feature for for English, and the 26A feature for Dutch, as well as all features for all non-Germanic languages.
The model is then evaluated on the held-out features for Germanic highlighted in red, i.e., 26A for English, 81A for Dutch, and both of these for German.
This is important, as knowing that a language is SVO would make it trivial to infer that it is not, e.g., SOV.

\paragraph{Unlabelled multilingual data.}
To induce language embeddings, we need a considerable amount of multilingual unlabelled data.
We use an in-house version of the massively multilingual Bible corpus, so as to have comparable data for all languages, although parallel data is not a strict necessity for our method.\footnote{Some languages available via \url{http://homepages.inf.ed.ac.uk/s0787820/bible/}}
We train the multilingual language model on a collection of 975 languages, each with approximately 200,000 tokens available.
We only train on languages for which the symbol size is relatively modest, a criterion which we fulfil by only using translations with Latin, Cyrillic, and Greek alphabets.\looseness=-1

Language used in the bible differs substantially from most modern language use, which would be a challenge if one were interested in transferring the language model itself.
Here, we are only interested in the distributed language embeddings for each language $\ell$.
It is safe to assume that the typological features underlying the texts we use will be representative of those coded in WALS, hence the domain should not matter much and the method should work equally well given any domain of input texts for the unsupervised training of language embeddings.

\section{Experiments}

\subsection{General Experimental Setup}

As described in \cref{sec:data}, we binarise WALS.
In order to compare directly with our semi-supervised extension, we limit our evaluation to the subset of languages which is the intersection of the languages for which we have Bible data, and the languages which are present in WALS.
Finally, we observe that some languages have very few features encoded, and some features are encoded for very few languages.
For instance, feature 10B (Nasal Vowels in West Africa), is only encoded for a total of 40 languages, and only one feature value appears for more than 10 languages.
Because of this, we restrict our evaluation to languages and feature values which occur at least 10 times.
Note that we evaluate on the original parameters, and not the binarised ones.\looseness=-1

Our general experimental set-up is as follows.
We first
split the languages in WALS into each language branch (\textit{genus} using WALS terminology).
This gives us, e.g., a set of Germanic languages, a set of Romance languages, a set of Berber languages, and so on. (We note that this does not correspond to the notion of a language family, e.g., the Indo-European language family.)
We wish to evaluate on this type of held-out set, as it is both relatively challenging: If we know the parameters of Portuguese, predicting the parameters for Spanish is a much easier task.
This setup will both give us a critical estimate of how well we can predict features overall, in addition to mimicking a scenario in which we either have a poorly covered language or branch, which we wish to add to WALS.

\begin{figure}[tbp]
	\includegraphics[width=0.9\columnwidth]{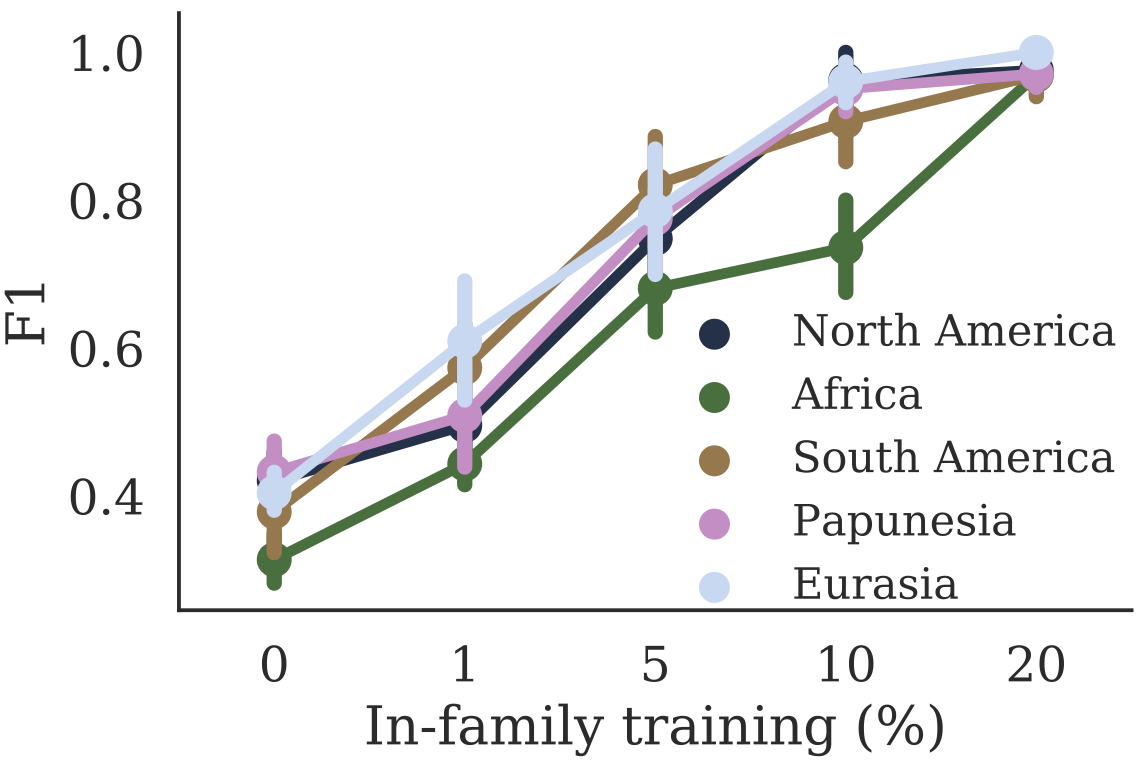}
    \caption{\label{fig:macroareas} F1 score for feature prediction separated by macroarea with varying in-branch training data.}
\end{figure}

\subsection{Typological Collaborative Filtering}

We evaluate our method for typological collaborative filtering on each language branch in WALS in a series of experiments.
Given a branch $B$ (a set of languages), we randomly select 80\% of the feature-language combinations from the languages $\ell\in B$, which we use for evaluation (e.g.~those highlighted in red in \cref{fig:matrix}).
The remaining 20\% is either not considered, or (partially) used for training, as we run experiments in which we train on $(0,1,5,10,20)$\% relative of the held-out data.
The idea behind this is that it should be very difficult to predict features if nothing is known for a language at all, whereas knowing a few features of a language, or of related languages, should allow the model to take advantage of the strong correlations between features (\cref{fig:corr-heat}) and between languages (\cref{fig:corr-langs}). 
We train on all data in WALS for languages which are not in the current evaluation branch under consideration.

Each experiment is then an evaluation of how well we can predict features for a completely or relatively unseen language family.
Evaluation is done across the branches in WALS with more than four languages represented, after filtering away languages for which we have fewer than 10 features available.
This amounts to a total of 36 branches, and 448 languages.
We repeat each experiment 5 times per language branch, for each proportion of in-branch training data in $(0,1,5,10,20)$\%, yielding a total of 900 experiment runs.
The results reported are the mean across these runs.

\cref{fig:macroareas} shows the micro F1 score we obtain averaged across macroareas.
The bars indicate 95\% confidence intervals.
We can see that, with access to 20\% of the in-branch training data, we can predict features at above 90\% F1 score regardless of macroarea. 
Prediction generally is more challenging for languages in the macroarea Africa.
This can be explained by, e.g., contrasting with the Eurasian macroarea.
Whereas the latter includes branches which are relatively uncontroversial, such as Germanic and Slavic languages, this is not the case with the former.
One such example is Bongo-Bagirmi (one of the evaluation branches, spoken in Central Africa), for which there is poor agreement in terms of classification \citep{bender:2000}.

\subsection{Semi-supervised extension}

We next evaluate the semi-supervised extension, which requires unlabelled texts for a large amount of languages,  although the domain of these texts should not matter much.
This allows us to take advantage of correlations between similar languages to point the model in the right direction. 
Even with 1\% training data, it may be very useful for a model to know that, e.g., German and Dutch are grammatically very similar.
Hence, if the 1\% training data contains features for Dutch, it should be quite easy for the model to learn to transfer these to German.

\begin{figure}[tbp]
	\includegraphics[width=0.9\columnwidth]{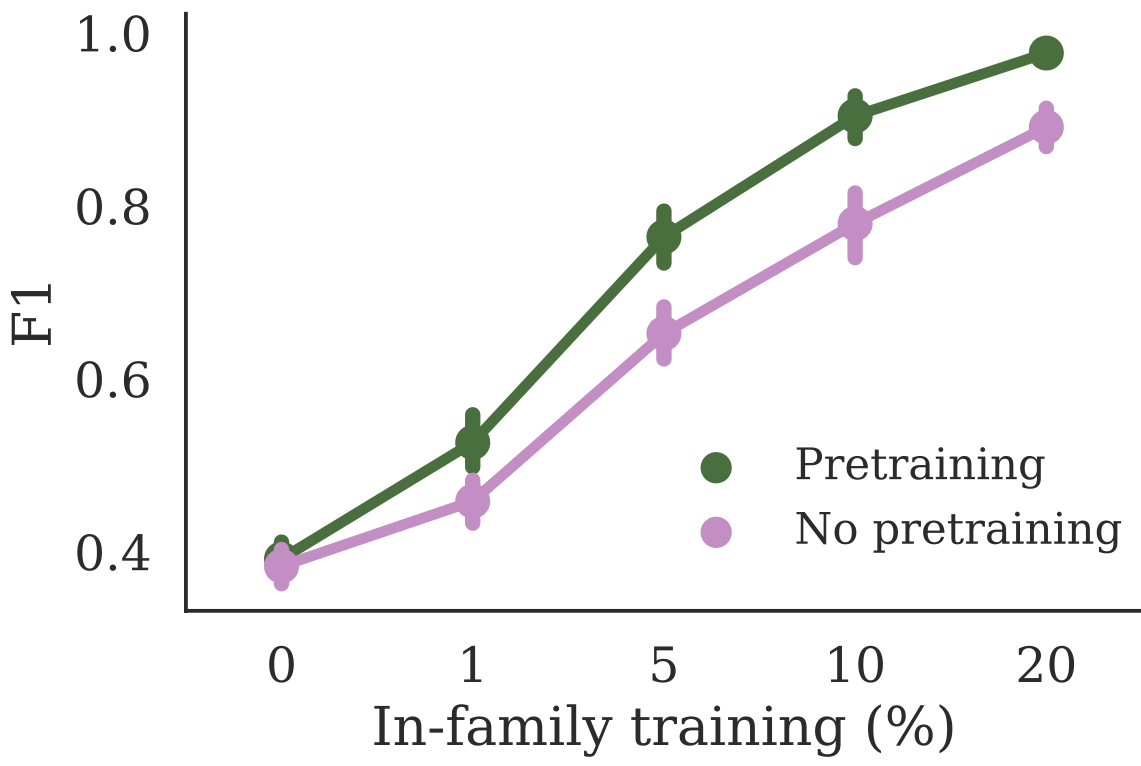}
    \caption{\label{fig:pretraining} F1 scores across all held-out language families, comparing pretraining with no pretraining.}
\end{figure}

\cref{fig:pretraining} shows results with pre-training. 
Without any in-branch training data, using pretrained embeddings does not offer any improvement in prediction power. 
This can be explained by the fact % that we are updating our 
language embeddings are updated during training, which leads to a drift in the representations present in the training material. 
This was chosen since not updating the representations yielded poor performance in all settings explored. 
We hypothesise that this is because, although a language modelling objective offers a good starting point in terms of encoding typological features, it is not sufficient to explain the typological diversity of languages.
For instance, a language model should hardly care about the phonological nature of a language.
This is in line with previous work which shows that the linguistic nature of the target task is important when predicting typological features with language embeddings \citep{bjerva_augenstein:naacl:18}. 
However, once we have access to $>5\%$ of in-branch training data, language embeddings offers a substantial improvement, e.g.~an F1 error reduction of more than 50\% with access to 10\% of the in-branch data (see \cref{tab:resultsAll} and \cref{tab:results} for per-branch results in the appendix).  
This shows that we can partially aid a typologist's work by utilising unannotated data.\looseness=-1

\subsection{Quantitative comparison}
In addition to evaluating our method of typological CF, we compare to some baselines drawn from earlier work.
First, we report a most frequent value baseline.
As many typological features are heavily skewed, this is quite high already.
For instance, defaulting to the most frequent value for word order (i.e. SVO) would yield an accuracy of 41\% (\textbf{Freq.} in \cref{tab:resultsAll}).
A more involved baseline is \citet{bjerva_augenstein:naacl:18}, who use pre-trained language embeddings in a k-NN classifier trained on individual WALS features (\textbf{Individual pred.} in \cref{tab:resultsAll}).
For the baseline reported here, we only use one nearest neighbour for this prediction.
The scores we obtain here are quite low compared to \citet{bjerva_augenstein:naacl:18}, which is explained by the fact that we have access to very little training data in the current setting, and highlights the importance of taking advantage of correlations between languages and features, and not simply looking at these factors in isolation.
Finally, we compare our typological collaborative filtering approach, as well as our semi-supervised extension (\textbf{T-CF} and \textbf{SemiSup} in \cref{tab:resultsAll}).

\begin{table}
  \centering
  \resizebox{\columnwidth}{!}{
  \begin{tabular}{cccccc}
    \toprule
     \textbf{In-branch train \%} & \textbf{Freq. F1} & \textbf{Indiv. pred. F1} & \textbf{T-CF F1} &\textbf{SemiSup F1} \\
    \midrule
 0.00 & 0.2950 & 0.2990 & 0.3998 &  0.3916 \\
 0.01 & 0.2949 & 0.2976 & 0.4578 &  0.5263 \\
 0.05 & 0.2947 & 0.2970 & 0.6552 &  0.7641 \\
 0.10 & 0.2945 & 0.2971 & 0.7807 &  0.9040 \\
 0.20 & 0.2938 & 0.2973 & 0.8835 &  \textbf{0.9767} \\
    \bottomrule
  \end{tabular}
  }
  \caption{\label{tab:resultsAll} Aggregate results w. 0-20\% relative use of in-branch training data. Columns show: most frequent class (Freq.), individual prediction per feature with language-embeddings (Individual pred.), typological collaborative filtering (T-CF), semi-supervised extension (SemiSup). For standard deviations across runs and per-branch results, see appendix.}
\end{table}

\section{Analysis}

Accuracy for several experimental settings is visualised in  Figure~\ref{fig:accuracy}, broken down by the linguistic category of the predicted features. Since results change little between the 5\% in-branch setting and higher percentages, we only look at 0\%, 1\% and 5\% here. We also visualise accuracy without  (\cref{fig:accuracy}, left) and with our semi-supervised extension (\cref{fig:accuracy}, right) in each setting. 

\subsection{Typological collaborative filtering with variable amounts of data}

Focussing on \cref{fig:accuracy} (left) alone first we observe an expected pattern: using increasingly more in-branch data boosts performance across all feature groups. This increase in accuracy can be attributed to the model having more knowledge about each query language in itself and about how languages relate to one another, based on similarities in their parameter configurations. 

Making a prediction about the order of adposition and noun phrase in a given language, lacking any other information about that language, is basically a shot in the dark. In-branch training data, in our experiments, includes in-language training data, too. Having one piece of information about the word order of that language, its ordering of relative clause and noun, or even its affixational properties, immediately makes the prediction informed rather than random: in many other languages within and outside this particular language family the model would have likely observed a strong correlation between these features and the order of adposition and noun phrase, which are all subject to the more general headedness parameter.

Certain features and feature configurations may not be as abundant cross-linguistically as the set of headedness features. In those cases, access to in-branch data is crucial. Consider e.g. the feature 10B Nasal Vowels in West Africa: a handful of language branches exhibit this feature and at least one of its values, \textit{no nasal vs. oral vowel contrast}, is characteristic predominantly of Niger-Congo languages. Without \textit{any} in-branch training data, the model's knowledge of this feature value is extremely limited, making its correct prediction for a Niger-Congo language virtually impossible. A small amount of in-branch training data thus increases the chance of a correct prediction greatly.

\subsection{Semi-supervised extension
}
\begin{figure}[t]
	\centering
	\includegraphics[width=\columnwidth]{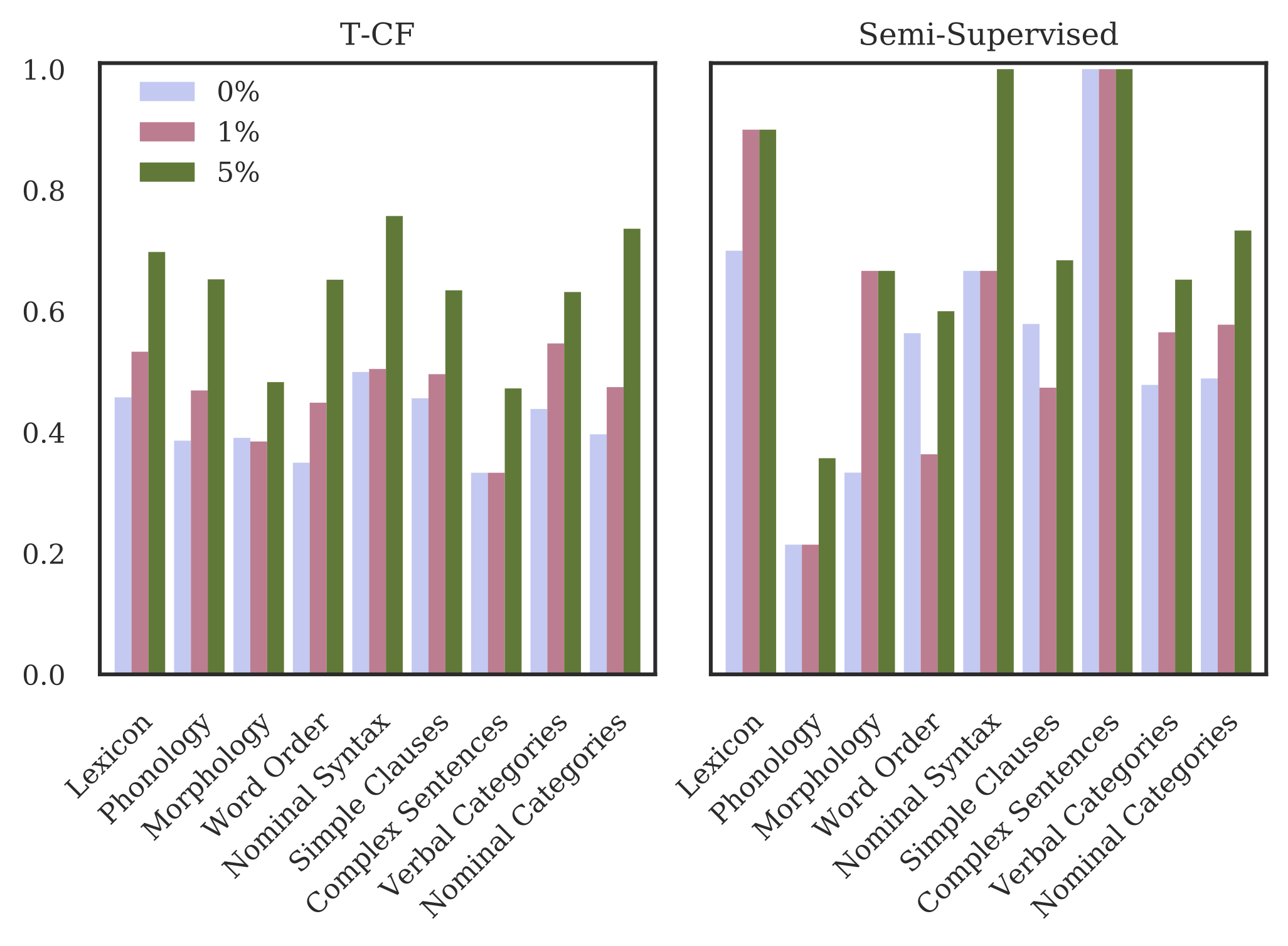}
    \caption{\label{fig:accuracy} Accuracy per feature group (Germanic).}
\end{figure}

Comparing \cref{fig:accuracy} to \cref{fig:pretraining} reveals a crucial finding. While we see very little improvement from pretraining for 0\% in-branch training overall, for individual linguistic categories, it mostly benefits prediction: seven out of nine feature groups are predicted more accurately with pretraining. Phonological and morphological predictions experience moderate deterioration, however, counterbalancing much of the improvement in other categories, which leads to the overall result of seemingly little improvement from pretraining. The limited effect of pretraining on prediction of phonological and morphological features can be explained with reference to the richness and complexity of these linguistic domains, which makes for data sparsity and generally makes them harder to learn based on distributional information alone. Moreover, a number of phonological features refer to aspects of language that may not be reflected in writing, such as stress and devoicing. All other categories concern syntactic and semantic information, which is known to be learnable from word distribution, and therefore benefit from the knowledge carried by language embeddings.

\cref{fig:accuracy} (right) shows an unsteady interaction between pretraining and the addition of increasing amounts of in-branch data. While pretraining alone helps for predicting most features, as pointed out above, an extra 1\% of in-branch data in the pretrained setting has a rather variable impact across feature groups. For a few groups it helps, as is expected, for a few it has no effect and for two groups, `Word Order' and `Simple Clauses', it makes for quite a drop in accuracy. We speculate that this effect, while negative, is indicative of the general power of language embeddings in associating related languages. Consider the test query `Fixed Stress Location' in English, where the 1\% of in-branch training data contains the information in Table~\ref{tab:traindata}. Based on feature correlation alone, the model should predict `No fixed stress' for English, since this value always co-occurs with `Right-oriented stress'. Yet, due to the proximity in the English and Icelandic embeddings, the model may copy the value of Icelandic and falsely predict `Initial stress' for English, too.
The risk of this happening decreases with more in-branch training data, since the model can generalise over more in-branch features.

\begin{table}%[htbp]
  \centering
  \resizebox{\linewidth}{!}{
  \begin{tabular}{lrrr}
    \toprule
    \textbf{Language} & \textbf{Genus} &  \textbf{Fixed Stress Location} & \textbf{Weight-Sensitive Stress} \\
    \midrule
    English&Germanic&?&Right-oriented\\
    Icelandic&Germanic&Initial&Fixed stress
\\
    \midrule
  \end{tabular}
  }
  \caption{\label{tab:traindata} In-branch training data in example scenario}
\end{table}

Lastly, notice that accuracy for phonological features remains low even with 5\% of in-branch data, and it is lower in the pretrained setting compared to the no-pretraining one. 
This brings us to the conclusion that using pretrained embeddings which are fine-tuned for specific tasks which encode different linguistic levels, as in \citet{bjerva_augenstein:naacl:18}, might also be useful in our semi-supervised extension of typological collaborative filtering.

\section{Related Work}\label{sec:RelWork}

\paragraph{Computational Typology} 
The availability of unlabelled datasets for hundreds of languages permits inferring linguistic properties and categories \citep{ostling:2015,asgari:typology}.
Individual prediction of typological features has been attempted in conjunction with several NLP tasks \citep{malaviya:2017,bjerva_augenstein:naacl:18,bjerva:2017:uralic}.
Our work is most similar to \citet{murawaki:2017}, who presents a Bayesian approach to utilising relations between features and languages for feature prediction. 
However, our work differs on several important counts, as we
(i) include language information obtained through unsupervised learning, which allows us to take advantage of raw data and predict features for completely unannotated languages, 
(ii) analyse the effects of varying amounts of known features, especially in situations with and without in-branch training data,
and (iii) view the problem of typological features through the lens of parameters from principles and parameters \citep{chomsky_smith_2000}.
Deep generative models have also been explored previously for modelling phonology \citep{cotterell:2017}.
Our work builds on these research directions, by (i) developing a deep generative model which (ii) takes advantage of correlations, rather than predicting features individually, and (iii) exploits unlabelled data.
This work is also related to linguistic representations encoded in neural models \citep{kadar:2017} and language embeddings \citep{bjerva:cl}, multilingual relations between languages in various representational levels \citep{beinborn2019semantic}, as well as the related problem of phylogenetic inference \citep{farach:95,nichols:08}.
For a survey of typology in NLP, see \citet{ponti:2018}.

\paragraph{Matrix Factorisation}

Collaborative Filtering was popularised in the early 1990s as a technique for recommender systems with applications such as mail filtering \cite{Goldberg92}, and article \cite{GroupLens94} and movie recommendation \cite{MovieLens98}. %Machine learning 
Model-based algorithms soon became popular \cite{breese1998} to overcome the cold start problem arising for unseen users or items at test time. The most successful one of these, in turn, is matrix factorisation, as applied in this paper, which represents users and items as (dense) vectors in the same latent feature space and measures their compatibility by taking the dot product between the two representations \cite{10.1109/MC.2009.263,bokde2015matrix}. 
Beyond recommender systems, matrix factorisation has shown successes in a wide variety of subareas of NLP \cite{Riedel2013,conf/naacl/RocktaschelSR15,levy2014neural,lei2014lowrank,conf/naacl/AugensteinRS18}.

\section{Conclusion}
We introduce a generative model inspired by the principles-and-parameters framework, drawing on the correlations between typological features of languages to solve the novel task of typological collaborative filtering.
We further show that raw text can be utilised to infer similarities between languages, thus allowing for extending the method with semi-supervised language embeddings.

\section*{Acknowledgements}
We acknowledge the computational resources provided by CSC in Helsinki through NeIC-NLPL (www.nlpl.eu), and the support of the NVIDIA Corporation with the donation of the Titan Xp GPU used for this research. The third author acknowledges support from a Facebook Fellowship.

\bibliography{wals}

\begin{thebibliography}{46}
\expandafter\ifx\csname natexlab\endcsname\relax\def\natexlab#1{#1}\fi

\bibitem[{Asgari and Sch{\"u}tze(2017)}]{asgari:typology}
Ehsaneddin Asgari and Hinrich Sch{\"u}tze. 2017.
\newblock \href {https://doi.org/10.18653/v1/D17-1011} {Past, present, future:
  A computational investigation of the typology of tense in 1000 languages}.
\newblock In \emph{Proceedings of the 2017 Conference on Empirical Methods in
  Natural Language Processing}, pages 113--124, Copenhagen, Denmark.
  Association for Computational Linguistics.

\bibitem[{Augenstein et~al.(2018)Augenstein, Ruder, and
  S{\o}gaard}]{conf/naacl/AugensteinRS18}
Isabelle Augenstein, Sebastian Ruder, and Anders S{\o}gaard. 2018.
\newblock \href {https://doi.org/10.18653/v1/N18-1172} {Multi-task learning of
  pairwise sequence classification tasks over disparate label spaces}.
\newblock In \emph{Proceedings of the 2018 Conference of the North American
  Chapter of the Association for Computational Linguistics: Human Language
  Technologies, Volume 1 (Long Papers)}, pages 1896--1906, New Orleans,
  Louisiana. Association for Computational Linguistics.

\bibitem[{Baker(2008)}]{baker}
Mark~C Baker. 2008.
\newblock The macroparameter in a microparametric world.
\newblock \emph{The Limits of Syntactic Variation}, 132.

\bibitem[{Beinborn and Choenni(2019)}]{beinborn2019semantic}
Lisa Beinborn and Rochelle Choenni. 2019.
\newblock Semantic drift in multilingual representations.
\newblock \emph{arXiv preprint arXiv:1904.10820}.

\bibitem[{Bender(2000)}]{bender:2000}
M~Lionel Bender. 2000.
\newblock Nilo-{S}aharan.
\newblock In B.~Heine and Nurse. D., editors, \emph{African languages: An
  introduction}, pages 43--73. Cambridge University Press Cambridge.

\bibitem[{Biberauer et~al.(2009)Biberauer, Holmberg, Roberts, and
  Sheehan}]{biberauer}
Theresa Biberauer, Anders Holmberg, Ian Roberts, and Michelle Sheehan. 2009.
\newblock \emph{Parametric variation: Null subjects in minimalist theory}.
\newblock Cambridge University Press.

\bibitem[{Bjerva and
  Augenstein(2018{\natexlab{a}})}]{bjerva_augenstein:naacl:18}
Johannes Bjerva and Isabelle Augenstein. 2018{\natexlab{a}}.
\newblock \href {https://doi.org/10.18653/v1/N18-1083} {From phonology to
  syntax: Unsupervised linguistic typology at different levels with language
  embeddings}.
\newblock In \emph{Proceedings of the 2018 Conference of the North American
  Chapter of the Association for Computational Linguistics: Human Language
  Technologies, Volume 1 (Long Papers)}, pages 907--916, New Orleans,
  Louisiana. Association for Computational Linguistics.

\bibitem[{Bjerva and Augenstein(2018{\natexlab{b}})}]{bjerva:2017:uralic}
Johannes Bjerva and Isabelle Augenstein. 2018{\natexlab{b}}.
\newblock \href {https://doi.org/10.18653/v1/W18-0207} {Tracking typological
  traits of uralic languages in distributed language representations}.
\newblock In \emph{Proceedings of the Fourth International Workshop on
  Computatinal Linguistics of Uralic Languages}, pages 76--86, Helsinki,
  Finland. Association for Computational Linguistics.

\bibitem[{Bjerva et~al.(2019)Bjerva, Östling, Han~Veiga, Tiedemann, and
  Augenstein}]{bjerva:cl}
Johannes Bjerva, Robert Östling, Maria Han~Veiga, Jörg Tiedemann, and
  Isabelle Augenstein. 2019.
\newblock {What do Language Representations Really Represent?}
\newblock \emph{Computational Linguistics}.
\newblock In press, arXiv preprint arXiv:1901.02646.

\bibitem[{Bokde et~al.(2015)Bokde, Girase, and Mukhopadhyay}]{bokde2015matrix}
Dheeraj Bokde, Sheetal Girase, and Debajyoti Mukhopadhyay. 2015.
\newblock {Matrix Factorization Model in Collaborative Filtering Algorithms: A
  Survey}.
\newblock \emph{Procedia Computer Science}, (49):136--146.

\bibitem[{Breese et~al.(1998)Breese, Heckerman, and Kadie}]{breese1998}
John~S. Breese, David Heckerman, and Carl Kadie. 1998.
\newblock Empirical analysis of predictive algorithms for collaborative
  filtering.
\newblock In \emph{14th Conference on Uncertainty in Artificial Intelligence},
  pages 43--52.

\bibitem[{Carnie(2013)}]{carnie2013syntax}
Andrew Carnie. 2013.
\newblock \emph{Syntax: {A} generative introduction}.
\newblock John Wiley \& Sons.

\bibitem[{Chomsky(1957)}]{chomsky2002syntactic}
Noam Chomsky. 1957.
\newblock \emph{Syntactic structures}.
\newblock Walter de Gruyter.

\bibitem[{Chomsky(1965)}]{chomsky1965}
Noam Chomsky. 1965.
\newblock \emph{Aspects of the Theory of Syntax}.
\newblock The MIT Press.

\bibitem[{Chomsky(1981)}]{chomsky1981lectures}
Noam Chomsky. 1981.
\newblock \emph{Lectures on government and binding: {T}he {P}isa lectures}.
\newblock 9. Walter de Gruyter.

\bibitem[{Chomsky(2000)}]{chomsky_smith_2000}
Noam Chomsky. 2000.
\newblock \emph{New Horizons in the Study of Language and Mind}.
\newblock Cambridge University Press.

\bibitem[{Collins et~al.(2001)Collins, Dasgupta, and
  Schapire}]{DBLP:conf/nips/CollinsDS01}
Michael Collins, Sanjoy Dasgupta, and Robert~E. Schapire. 2001.
\newblock A generalization of principal components analysis to the exponential
  family.
\newblock In \emph{{NIPS}}, pages 617--624.

\bibitem[{Cotterell and Eisner(2017)}]{cotterell:2017}
Ryan Cotterell and Jason Eisner. 2017.
\newblock \href {https://doi.org/10.18653/v1/P17-1109} {Probabilistic typology:
  Deep generative models of vowel inventories}.
\newblock In \emph{Proceedings of the 55th Annual Meeting of the Association
  for Computational Linguistics (Volume 1: Long Papers)}, pages 1182--1192,
  Vancouver, Canada. Association for Computational Linguistics.

\bibitem[{Croft(2002)}]{croft2002typology}
William Croft. 2002.
\newblock \emph{Typology and Universals}.
\newblock Cambridge University Press.

\bibitem[{Dahlen et~al.(1998)Dahlen, Konstan, Herlocker, Good, Borchers, and
  Riedl}]{MovieLens98}
B.J. Dahlen, J.A. Konstan, J.L. Herlocker, N.~Good, A.~Borchers, and J.~Riedl.
  1998.
\newblock {Jump-starting movieLens: User benefits of starting a collaborative
  filtering system with "dead-data"}.
\newblock In \emph{{ }}. University of Minnesota TR 98-017.

\bibitem[{Dryer and Haspelmath(2013)}]{wals}
Matthew~S. Dryer and Martin Haspelmath, editors. 2013.
\newblock \href {http://wals.info/} {\emph{WALS Online}}.
\newblock Max Planck Institute for Evolutionary Anthropology, Leipzig.

\bibitem[{Eisner(2002)}]{eisner-2002-cogsci}
Jason Eisner. 2002.
\newblock Discovering syntactic deep structure via {B}ayesian statistics.
\newblock \emph{Cognitive Science}, 26(3):255--268.

\bibitem[{Farach et~al.(1995)Farach, Kannan, and Warnow}]{farach:95}
Martin Farach, Sampath Kannan, and Tandy Warnow. 1995.
\newblock A robust model for finding optimal evolutionary trees.
\newblock \emph{Algorithmica}, 13:155--179.

\bibitem[{Goldberg et~al.(1992)Goldberg, Nichols, Oki, and Terry}]{Goldberg92}
David Goldberg, David Nichols, Brian Oki, and Douglas Terry. 1992.
\newblock \href {http://www.xerox.com/PARC/dlbx/tapestry-papers/TN44.ps} {Using
  collaborative filtering to weave an information tapestry}.
\newblock \emph{Communications of the ACM}, 35(12):61--70.
\newblock Special Issue on Information Filtering.

\bibitem[{Harris(1995)}]{harris1995linguistics}
Randy~Allen Harris. 1995.
\newblock \emph{The Linguistics Wars}.
\newblock Oxford University Press.

\bibitem[{Haspelmath(2008)}]{haspelmath:2008}
Martin Haspelmath. 2008.
\newblock Parametric versus functional explanations of syntactic universals.
\newblock \emph{The limits of syntactic variation}, 132:75--107.

\bibitem[{Hochreiter and Schmidhuber(1997)}]{lstm}
Sepp Hochreiter and J{\"u}rgen Schmidhuber. 1997.
\newblock Long short-term memory.
\newblock \emph{Neural computation}, 9(8):1735--1780.

\bibitem[{Kingma and Ba(2014)}]{adam}
Diederik Kingma and Jimmy Ba. 2014.
\newblock Adam: A method for stochastic optimization.
\newblock \emph{arXiv preprint arXiv:1412.6980}.

\bibitem[{Koren et~al.(2009)Koren, Bell, and Volinsky}]{10.1109/MC.2009.263}
Yehuda Koren, Robert Bell, and Chris Volinsky. 2009.
\newblock Matrix factorization techniques for recommender systems.
\newblock \emph{Computer}, 42(8):30--37.

\bibitem[{Kádár et~al.(2017)Kádár, Chrupała, and Alishahi}]{kadar:2017}
Ákos Kádár, Grzegorz Chrupała, and Afra Alishahi. 2017.
\newblock Representation of linguistic form and function in recurrent neural
  network.
\newblock \emph{Computational Linguistics}, 43:761--780.

\bibitem[{Lei et~al.(2014)Lei, Xin, Zhang, Barzilay, and
  Jaakkola}]{lei2014lowrank}
Tao Lei, Yu~Xin, Yuan Zhang, Regina Barzilay, and Tommi Jaakkola. 2014.
\newblock \href {https://doi.org/10.3115/v1/P14-1130} {Low-rank tensors for
  scoring dependency structures}.
\newblock In \emph{Proceedings of the 52nd Annual Meeting of the Association
  for Computational Linguistics (Volume 1: Long Papers)}, pages 1381--1391,
  Baltimore, Maryland. Association for Computational Linguistics.

\bibitem[{Levy and Goldberg(2014)}]{levy2014neural}
Omer Levy and Yoav Goldberg. 2014.
\newblock {Neural Word Embedding as Implicit Matrix Factorization}.
\newblock In \emph{Advances in Neural Information Processing Systems 27}, pages
  2177--2185. Curran Associates, Inc.

\bibitem[{Malaviya et~al.(2017)Malaviya, Neubig, and Littell}]{malaviya:2017}
Chaitanya Malaviya, Graham Neubig, and Patrick Littell. 2017.
\newblock \href {https://doi.org/10.18653/v1/D17-1268} {Learning language
  representations for typology prediction}.
\newblock In \emph{Proceedings of the 2017 Conference on Empirical Methods in
  Natural Language Processing}, pages 2529--2535, Copenhagen, Denmark.
  Association for Computational Linguistics.

\bibitem[{Min et~al.(2013)Min, Grishman, Wan, Wang, and
  Gondek}]{conf/naacl/MinGW0G13}
Bonan Min, Ralph Grishman, Li~Wan, Chang Wang, and David Gondek. 2013.
\newblock \href {https://www.aclweb.org/anthology/N13-1095} {Distant
  supervision for relation extraction with an incomplete knowledge base}.
\newblock In \emph{Proceedings of the 2013 Conference of the North American
  Chapter of the Association for Computational Linguistics: Human Language
  Technologies}, pages 777--782, Atlanta, Georgia. Association for
  Computational Linguistics.

\bibitem[{Murawaki(2017)}]{murawaki:2017}
Yugo Murawaki. 2017.
\newblock \href {https://www.aclweb.org/anthology/I17-1046} {Diachrony-aware
  induction of binary latent representations from typological features}.
\newblock In \emph{Proceedings of the Eighth International Joint Conference on
  Natural Language Processing (Volume 1: Long Papers)}, pages 451--461, Taipei,
  Taiwan. Asian Federation of Natural Language Processing.

\bibitem[{Nichols and Warnow(2008)}]{nichols:08}
Johanna Nichols and Tandy Warnow. 2008.
\newblock Tutorial on computational linguistic phylogeny.
\newblock \emph{Language and Linguistics Compass}, 2:760--820.

\bibitem[{Nicolis and Biberauer(2008)}]{nicolis}
Marco Nicolis and Theresa Biberauer. 2008.
\newblock The null subject parameter and correlating properties.
\newblock \emph{The limits of syntactic variation}, 132:271.

\bibitem[{van Oostendorp(2015)}]{van2015parametric}
Marc van Oostendorp. 2015.
\newblock Parameters in phonological analysis: stress.
\newblock In A.~Fabregas, J.~Mateu, and M.~Putnam, editors, \emph{Contemporary
  Linguistic Parameters}, Contemporary Studies in Linguistics, pages 234--258.
  London: Bloomsbury.

\bibitem[{{\"O}stling(2015)}]{ostling:2015}
Robert {\"O}stling. 2015.
\newblock \href {https://doi.org/10.3115/v1/P15-2034} {Word order typology
  through multilingual word alignment}.
\newblock In \emph{Proceedings of the 53rd Annual Meeting of the Association
  for Computational Linguistics and the 7th International Joint Conference on
  Natural Language Processing (Volume 2: Short Papers)}, pages 205--211,
  Beijing, China. Association for Computational Linguistics.

\bibitem[{{\"O}stling and Tiedemann(2017)}]{ostling_tiedemann:2017}
Robert {\"O}stling and J{\"o}rg Tiedemann. 2017.
\newblock \href {https://www.aclweb.org/anthology/E17-2102} {Continuous
  multilinguality with language vectors}.
\newblock In \emph{Proceedings of the 15th Conference of the European Chapter
  of the Association for Computational Linguistics: Volume 2, Short Papers},
  pages 644--649, Valencia, Spain. Association for Computational Linguistics.

\bibitem[{Ponti et~al.(2018)Ponti, O'Horan, Berzak, Vuli{\'c}, Reichart,
  Poibeau, Shutova, and Korhonen}]{ponti:2018}
Edoardo~Maria Ponti, Helen O'Horan, Yevgeni Berzak, Ivan Vuli{\'c}, Roi
  Reichart, Thierry Poibeau, Ekaterina Shutova, and Anna Korhonen. 2018.
\newblock Modeling language variation and universals: A survey on typological
  linguistics for natural language processing.
\newblock \emph{Computational Linguistics, arXiv preprint arXiv:1807.00914}.

\bibitem[{Prince and Smolensky(2008)}]{prince2008optimality}
Alan Prince and Paul Smolensky. 2008.
\newblock \emph{Optimality Theory: {C}onstraint interaction in generative
  grammar}.
\newblock John Wiley \& Sons.

\bibitem[{Resnick et~al.(1994)Resnick, Iacovou, Suchak, Bergstrom, and
  Riedl}]{GroupLens94}
P.~Resnick, N.~Iacovou, M.~Suchak, P.~Bergstrom, and J.~Riedl. 1994.
\newblock {GroupLens: An Open Architecture for Collaborative Filtering of
  Netnews}.
\newblock In \emph{{Proceedings of the 1994 Conference on Computer Supported
  Collaborative Work}}, pages 175--186.

\bibitem[{Riedel et~al.(2013)Riedel, Yao, McCallum, and Marlin}]{Riedel2013}
Sebastian Riedel, Limin Yao, Andrew McCallum, and Benjamin~M. Marlin. 2013.
\newblock \href {https://www.aclweb.org/anthology/N13-1008} {Relation
  extraction with matrix factorization and universal schemas}.
\newblock In \emph{Proceedings of the 2013 Conference of the North American
  Chapter of the Association for Computational Linguistics: Human Language
  Technologies}, pages 74--84, Atlanta, Georgia. Association for Computational
  Linguistics.

\bibitem[{Rockt{\"a}schel et~al.(2015)Rockt{\"a}schel, Singh, and
  Riedel}]{conf/naacl/RocktaschelSR15}
Tim Rockt{\"a}schel, Sameer Singh, and Sebastian Riedel. 2015.
\newblock \href {https://doi.org/10.3115/v1/N15-1118} {Injecting logical
  background knowledge into embeddings for relation extraction}.
\newblock In \emph{Proceedings of the 2015 Conference of the North American
  Chapter of the Association for Computational Linguistics: Human Language
  Technologies}, pages 1119--1129, Denver, Colorado. Association for
  Computational Linguistics.

\bibitem[{Tsvetkov et~al.(2016)Tsvetkov, Sitaram, Faruqui, Lample, Littell,
  Mortensen, Black, Levin, and Dyer}]{tsvetkov:2016}
Yulia Tsvetkov, Sunayana Sitaram, Manaal Faruqui, Guillaume Lample, Patrick
  Littell, David Mortensen, Alan~W. Black, Lori Levin, and Chris Dyer. 2016.
\newblock \href {https://doi.org/10.18653/v1/N16-1161} {Polyglot neural
  language models: A case study in cross-lingual phonetic representation
  learning}.
\newblock In \emph{Proceedings of the 2016 Conference of the North American
  Chapter of the Association for Computational Linguistics: Human Language
  Technologies}, pages 1357--1366, San Diego, California. Association for
  Computational Linguistics.

\end{thebibliography}
\bibliographystyle{acl_natbib}

\appendix

\section{Appendices}
\begin{table*}[!ht]
  \centering
  \resizebox{\textwidth}{!}{
  \begin{tabular}{llrrrrrr}
    \toprule
    \textbf{Macroarea} & \textbf{Branch} & \textbf{In-branch training (\%)} & \textbf{Freq. F1} & \textbf{Individual pred. F1} & \textbf{T-CF F1} &\textbf{SemiSup F1} \\
    \midrule
\multirow{5}{*}{Africa} & \multirow{5}{*}{Bantoid} & 0.00 & 0.3510 (0.0042) & 0.3540 (0.0057) & 0.3527 (0.0260) & 0.3287 (0.0292) \\
                          & & 0.01 & 0.3510 (0.0042) & 0.3643 (0.0077) & 0.4100 (0.0196) &  0.4013 (0.0260) \\
                         & & 0.05 & 0.3510 (0.0036) & 0.3523 (0.0053) & 0.4997 (0.0446) &  0.5467 (0.0266) \\
                         & & 0.10 & 0.3507 (0.0031) & 0.3570 (0.0029) & 0.5687 (0.1065) &  0.6107 (0.0963) \\
                         & & 0.20 & 0.3480 (0.0057) & 0.3450 (0.0071) & 0.8397 (0.0475) &  \textbf{0.8620} (0.0645) \\
                           \midrule
                          \multirow{5}{*}{Papunesia} &  \multirow{5}{*}{Chimbu} & 0.00 & 0.2860 (0.0182) & 0.2940 (0.0142) & 0.3460 (0.0771) & 0.3357 (0.0708) \\
                           & & 0.01 & 0.2860 (0.0182) & 0.2887 (0.0120) & 0.3600 (0.0805) &  0.2773 (0.0890) \\
                           & & 0.05 & 0.2850 (0.0171) & 0.2990 (0.0236) & 0.3833 (0.1012) &  0.4560 (0.2013) \\
                           & & 0.10 & 0.2850 (0.0171) & 0.2937 (0.0161) & 0.6140 (0.0884) &  0.6950 (0.0318) \\
                           & & 0.20 & 0.2873 (0.0181) & 0.2893 (0.0135) & 0.7800 (0.1181) &  \textbf{0.8267} (0.1605) \\
     \midrule
 \multirow{5}{*}{Africa} & \multirow{5}{*}{Kwa} & 0.00 & 0.2417 (0.0111) & 0.2540 (0.0099) & 0.3163 (0.0381) & 0.2720 (0.0490) \\
                           & & 0.01 & 0.2407 (0.0104) & 0.2550 (0.0051) & 0.3807 (0.0775) &  0.3843 (0.0796) \\
                           & & 0.05 & 0.2417 (0.0111) & 0.2597 (0.0095) & 0.4927 (0.1181) &  0.5290 (0.0978) \\
                           & & 0.10 & 0.2417 (0.0111) & 0.2453 (0.0076) & 0.6827 (0.1092) &  0.6827 (0.1092) \\
                           & & 0.20 & 0.2417 (0.0111) & 0.2443 (0.0033) & \textbf{0.9960} (0.0057) &  \textbf{0.9960} (0.0057) \\
    \midrule
\multirow{5}{*}{North America} & \multirow{5}{*}{Mixtecan} & 0.00 & 0.3507 (0.0076) & 0.3427 (0.0047) & 0.4387 (0.1244) & 0.4747 (0.0766) \\
                          & & 0.01 & 0.3507 (0.0076) & 0.3360 (0.0065) & 0.4593 (0.0514) &  0.5010 (0.0304) \\
                          & & 0.05 & 0.3500 (0.0083) & 0.3460 (0.0065) & 0.6867 (0.2274) &  0.7803 (0.0988) \\
                          & & 0.10 & 0.3490 (0.0094) & 0.3400 (0.0114) & 0.8610 (0.1288) &  0.8830 (0.1441) \\
                          & & 0.20 & 0.3490 (0.0094) & 0.3383 (0.0130) & 0.9883 (0.0165) &  \textbf{1.0000} (0.0000) \\
    \midrule
\multirow{5}{*}{Papunesia} & \multirow{5}{*}{Oceanic} & 0.00 & 0.3837 (0.0048) & 0.4030 (0.0096) & 0.3677 (0.0416) & 0.3663 (0.0433) \\
                          & & 0.01 & 0.3837 (0.0048) & 0.3957 (0.0066) & 0.3783 (0.0324) &  0.4047 (0.0217) \\
                          & & 0.05 & 0.3833 (0.0041) & 0.3993 (0.0068) & 0.6193 (0.0382) &  0.6753 (0.0838) \\
                          & & 0.10 & 0.3830 (0.0051) & 0.3983 (0.0071) & 0.7243 (0.1116) &  0.7337 (0.1245) \\
                          & & 0.20 & 0.3817 (0.0034) & 0.4057 (0.0113) & 0.8393 (0.0662) &  \textbf{0.9200} (0.0641) \\
                          \midrule
\multirow{5}{*}{All} & \multirow{5}{*}{All} & 0.00 & 0.2950 (0.0453) & 0.2990 (0.0448) & 0.3998 (0.1095) & 0.3916 (0.1077) \\
                          & & 0.01 & 0.2949 (0.0453) & 0.2976 (0.0457) & 0.4578 (0.1298) &  0.5263 (0.1605) \\
                          & & 0.05 & 0.2947 (0.0452) & 0.2970 (0.0436) & 0.6552 (0.1726) &  0.7641 (0.1572) \\
                          & & 0.10 & 0.2945 (0.0451) & 0.2971 (0.0434) & 0.7807 (0.1782) &  0.9040 (0.1324) \\
                          & & 0.20 & 0.2938 (0.0446) & 0.2973 (0.0451) & 0.8835 (0.1278) &  \textbf{0.9767} (0.0464) \\
    \bottomrule
  \end{tabular}
  }
  \caption{\label{tab:results} Aggregate results with 0-20\% relative use of in-branch training data. The columns indicate the most frequent class (Freq.), individual prediction per feature with language-embeddings (Individual pred.), typological collaborative filtering (T-CF), semi-supervised extension (SemiSup).}
\end{table*}

\end{document}